\documentclass[10pt,twocolumn,letterpaper]{article}

\usepackage[pagenumbers]{cvpr} 

\usepackage{graphicx}
\usepackage{amsmath}
\usepackage{amssymb}
\usepackage{booktabs}

\usepackage[pagebackref,breaklinks,colorlinks]{hyperref}

\usepackage[capitalize]{cleveref}
\crefname{section}{Sec.}{Secs.}
\Crefname{section}{Section}{Sections}
\Crefname{table}{Table}{Tables}
\crefname{table}{Tab.}{Tabs.}

\def\cvprPaperID{*****} 
\def\confName{CVPR}
\def\confYear{2024}

\begin{document}

\title{2nd Place Solution for PVUW Challenge 2024: Video Panoptic Segmentation}
\author{
Biao Wu\and
Diankai Zhang\and
Si Gao\and
Chengjian Zheng\and
Shaoli Liu\and
Ning Wang\and
State Key Laboratory of Mobile Network and Mobile Multimedia Technology,ZTE,China
\and
{\tt\small \{wu.biao,zhang.diankai,gao.si,zheng.chengjian,liu.shaoli,wangning\}@zte.com.cn}
}
\maketitle

\begin{abstract}
   Video  Panoptic  Segmentation (VPS) is a challenging task that is extends from image panoptic segmentation.VPS aims to simultaneously classify, track, segment all objects in a video, including both things and stuff. Due to its wide application in many downstream tasks such as video understanding, video editing, and autonomous driving. In order to deal with the task of video panoptic segmentation in the wild, we propose a robust integrated video panoptic segmentation solution. We use DVIS++ framework as our baseline to generate the initial masks. Then,we add an additional image semantic segmentation model to further improve the performance of semantic classes.Finally, our method achieves state-of-the-art performance with a VPQ score of 56.36 and 57.12 in the development and test phases, respectively, and ultimately ranked 2nd in the VPS track of the PVUW Challenge  at CVPR2024. 
\end{abstract}


\section{Introduction}
\label{sec:intro}
Panoptic segmentation \cite{rw0001} integrates the tasks of semantic segmentation and instance segmentation, requiring that each pixel of an image must be assigned a semantic label and an unique instance id. Since its inception, numerous studies \cite{rw0002, rw0003, rw0004, rw0005} have introduced a variety of innovative approaches aimed at enhancing both the accuracy and efficiency of this task. Video Panoptic Segmentation, as a direct extension of panoptic segmentation to videos, endeavors to consistently segment and identify all object instances across all frames.Numerous endeavors have focused on adapting image-based panoptic segmentation models for the video domain. VPSNet \cite{vps2020} combined the temporal feature fusion module and object tracking branch with a single-frame panoptic segmentation network to obtain panoptic video results. Panoptic-DeepLab \cite{rw0008} is the first bottom-up and single-shot panoptic segmentation model,utilizing a dual-ASPP and dual-decoder architecture tailored for semantic and instance segmentation. ViP-DeepLab \cite{rw0009} extended Panoptic-DeepLab \cite{rw0008} to jointly perform video panoptic segmentation and monocular depth estimation to address the inverse projection problem in vision. Note the disadvantages of previous methods that require multiple separate networks and complex post-processing, MaX-DeepLab \cite{rw0010} directly predicted masks and classes with a mask transformer, removing the needs for many hand-designed priors.Slot-VPS \cite{rw0011} designed a pioneering end-to-end framework that simplifies the VPS task by using a unified representation called panoptic slots to encode both foreground instances and background semantics in a video. DVIS \cite{rw0012} introduces a novel referring tracker for precise long-term alignment and a temporal refiner that leverages this alignment to effectively utilize temporal information, leading to improved instance segmentation outcomes. The 1st Place Solution for the CVPR 2023 PVUW VPS Track \cite{rw0013} embraced DVIS's strategy of dividing the task into three independent sub-tasks and optimizing for optimal outcomes. DVIS++ \cite{zhang2023dvisplus} improved the tracking capability of DVIS \cite{rw0012} by introducing a denoising training strategy and contrastive learning. Video-kMaX \cite{rw0015} extends the image segmenter for cliplevel video segmentation, and employed clip-kMaX for efficient clip-level segmentation and HiLA-MB for robust cross-clip association with hierarchical matching, effectively addressing both short- and long-term object tracking challenges. MaXTron \cite{rw0016} integrated a mask transformer with trajectory attention to perform VPS, bolstering temporal coherence through its within-clip and cross-clip tracking modules. 

    In summary, the progression of VPS has introduced sophisticated frameworks that offer innovative strategies, enhancing accuracy, efficiency, and temporal consistency in the segmentation and tracking of objects throughout video frames. These innovations have markedly advanced the frontier of video comprehension and analytical capabilities.
\section{Our solution}
\label{sec:method}
In this section, we will introduce the implementation process of our method. In order to deal with the task of video panoptic segmentation
in the wild, we propose a robust integrated video panoptic segmentation solution. In this solution, we first introduce DVIS++\cite{zhang2023dvisplus} as the baseline of video
panoptic segmentation and then choose ViT-adapter\cite{chen2022vitadapter} as the semantic segmentation baseline, and correct the sequence of ’stuff’ class
objects and individual sequences with only one ’thing’ class
object in panoptic segmentation through model ensemble.

\subsection{Video Panoptic Segmentation}

\begin{figure*}[ht]
    \centering
    \includegraphics[width= 16cm]{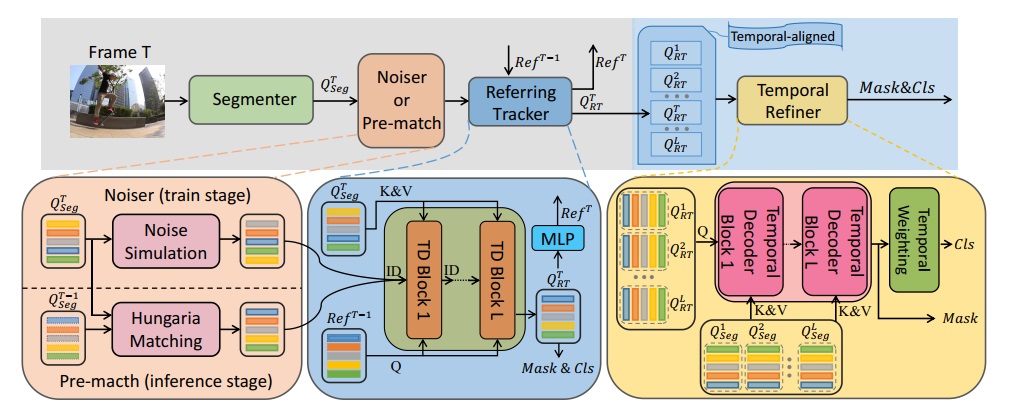}
    \caption{Architecture of DVIS++\cite{zhang2023dvisplus}.}
    \label{fig:my_diagram}
\end{figure*}

For video panoptic segmentation in the wild, DVIS++\cite{zhang2023dvisplus} is a decoupled video segmentation framework, which decouples video
segmentation into three cascaded sub-tasks: segmentation, tracking, and refinement, as shown in
Fig. 2. It is worth noting that unlike image segmentation, video segmentation involves capturing inter frame relationships from multiple frames for training. However, training consecutive frames requires a significant amount of GPU memory.To save memory resources, DVIS++ adopts a frozen DINOv2\cite{Dinov2} VIT backbone and employs Mask2Former\cite{cheng2022masked} as the segmenter, which is trained in three stages, sequentially training segmenter, referring tracker, and  temporal refiner.

\subsection{Video Semantic Segmentation}

Considering that VPSW and VIPSeg have the same data source and annotation category, and VPSW has a higher number of semantic segmentation annotation frames, which is very beneficial for training semantic segmentation models. In order to further improve the segmentation performance of stuff objects and some thing objects in panoptic segmentation, we introduce ViT-Adapter\cite{chen2022vitadapter} as a semantic segmentation baseline.

\begin{figure*}[ht]
    \centering
    \includegraphics[width= 16cm]{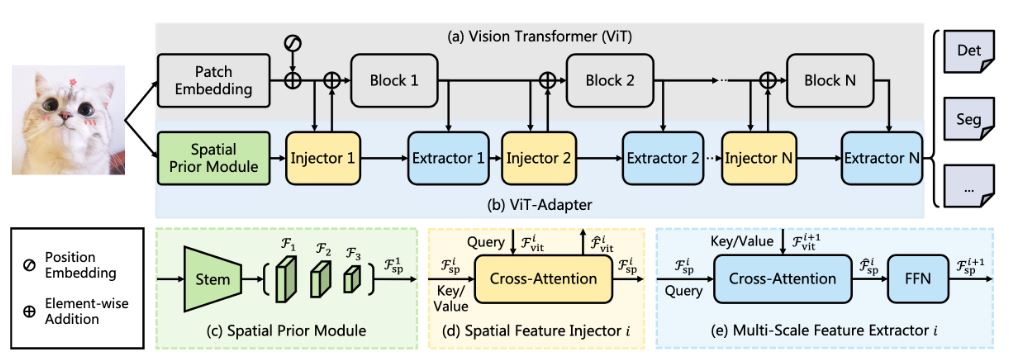}
    \caption{Architecture of ViT-Adapter\cite{chen2022vitadapter}.}
    \label{fig:my_diagram}
\end{figure*}

\begin{figure*}[ht]
    \centering
    \includegraphics[width= 16cm]{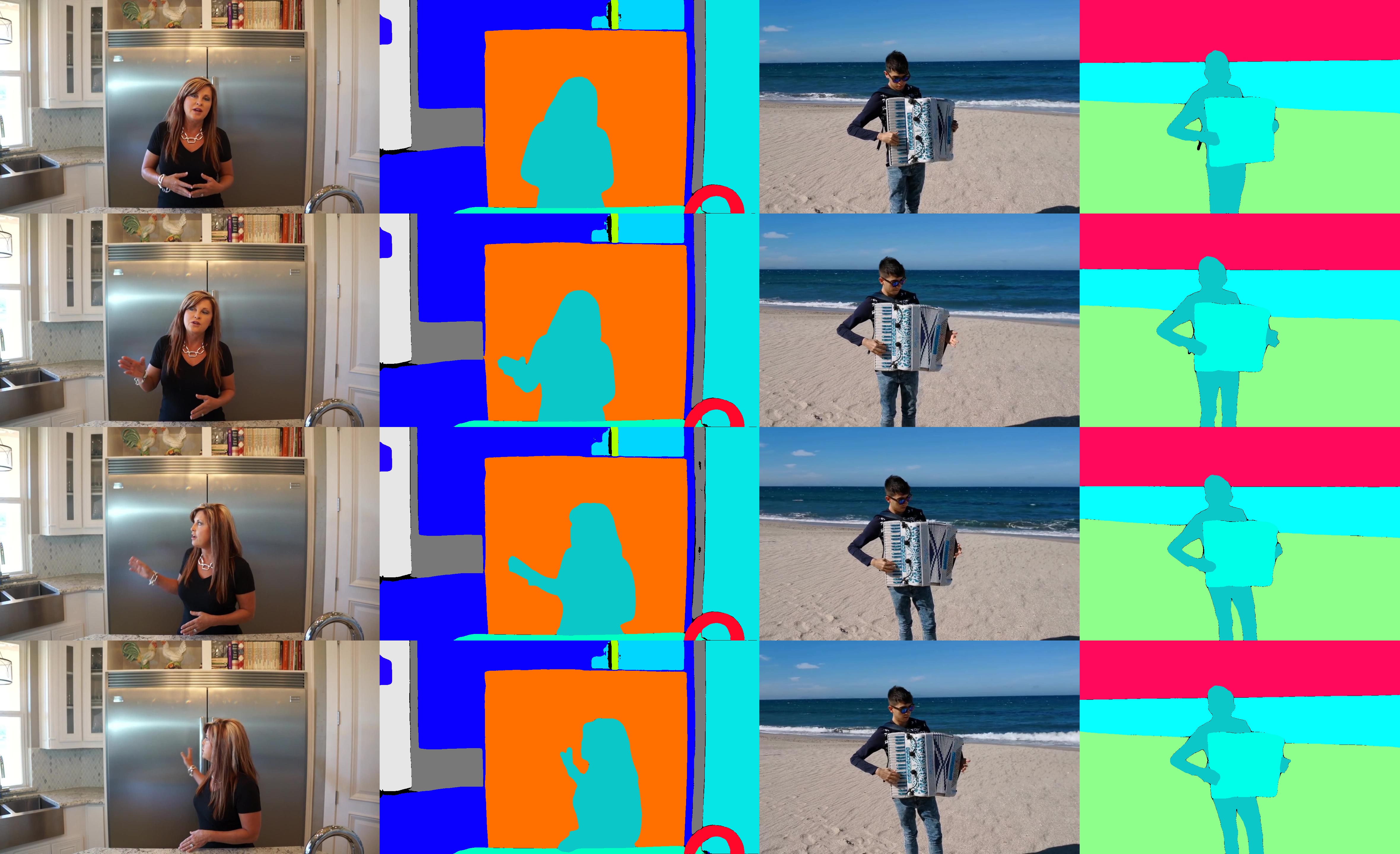}
    \caption{Qualitative result on VIPSeg test set of out method.}
    \label{fig:my_diagram}
\end{figure*}


\section{Experiments}
\label{sec:Experiments}

In this part, we will describe the implementation details of our proposed method and report the results on the PVUW2024  challenge test set.

\subsection{Datasets}
\textbf{VIPSeg.}\ VIPSeg \cite{miao2022large}provides 3,536 videos and 84,750 frames
with pixel-level panoptic annotations, covering a wide
range of real-world scenarios and categories, which is the
first attempt to tackle the challenging video panoptic segmentation task in the wild by considering diverse scenarios.
The train set, validation set, and test set of VIPSeg contain
2, 806/343/387 videos, respectively. 
VIPSeg showcases a variety of real-world scenes across 124 categories, consisting of 58 categories of 'thing' and 66 categories of 'stuff'. Due to limitations in computing resources, all the frames in VIPSeg are resized into 720P (the size of the short side is resized to 720)
for training and testing. 

\textbf{VSPW.} \ The VSPW\cite{miao2021vspw}  is a large-scale dataset for Video Semantic Segmentation, which is the first attempt to tackle the challenging
video scene parsing task in the wild by considering diverse
scenarios and annotates 124 categories of real-world scenarios, which contains 3,536 videos, with 251,633 frames totally. Among these videos, there are 2806 videos in the training set, 343 videos in the validation set, and 387 videos in the testing set. 

\subsection{Evaluation Metrics}
Video Panoptic Segmentation (VPS) Track of Pixel-level video understanding in the wild challenge uses VPQ\cite{vps2020} and STQ\cite{stq2021} to evaluate segmentation and tracking performance.Video Panoptic Quality (VPQ) for video panoptic
segmentation is based on PQ\cite{pq2019} (Panoptic Quality) and
computes the average quality by using tube IoU matching
across a small span of frames. Formally, the VPQ score across k frames is:

{
\small
\begin{equation}
\begin{aligned}
VPQ^k=\frac{1}{N_{c}}\sum_{c}p_{ij}(c)\frac{\sum_{p,g\in TP_c }IOU(p,g)}{|TP|_c^k+\frac{1}{2}|FP|_c^k + \frac{1}{2}|FN|_c^k}
\label{eq2}
\end{aligned}
\end{equation}
}

Segmentation and Tracking Quality (STQ) \cite{stq2021} is proposed to measure the segmentation quality and long tracking quality simultaneously.

For the 3nd Pixel-level Video Understanding in the Wild challenge(VPS Track), the ranking is evaluated according to VPQ.

\subsection{Implementation Details}

In our method, we employ ViT-L\cite{chen2022vitadapter} as the backbone and Mask2Former as the segmenter for video segmentation. We divide it into three stages to train the segmenter, referring tracker, and time refiner. In the first stage, we load the COCO\cite{coco} pre-trained weights to fine-tune the segmentation by using image level annotations from the training set of VIPSeg. In the second stage, we freeze the segmenter trained in the first stage and use a continuous 5-frame clip from the video as input. In the third stage, we only train the time refiner and freeze the segmenter and referring tracker trained in the first two stages, using continuous 21 frame clips as input.We train the  panoptic segmentation model on the training set of VIPSeg\cite{miao2022large} without using additional data such as validation set, conduct 40k iterations with a batch size of 4 and the learning rate is decayed by 0.1 at 26k iterations.
Multi-scale  training from 480 to 800 is used to randomly scale the short side of input video clips during training. Additionally, for training the refiner, we employ a random cropping strategy  with crop-size 608×608 from input video clips.For semantic segmentation, we use ViT- adapter\cite{chen2022vitadapter} as the baseline to train on the VSPW\cite{miao2021vspw} dataset.

\subsection{Ablation Studies}
For panoptic segmentation, we choose DVIS++ as the baseline and achieve a VPQ index of 55.93 in the test set stage. It shows good segmentation and tracking performance in handling 'stuff' and 'thing' objects. However, there are segmentation holes and category misjudgments in the output results of DVIS++, which seriously affected the VPQ score. In order to further improve the performance of the model, we choose ViT-adapter as the semantic segmentation baseline, and correct the sequence of 'stuff' class objects and individual sequences with only one 'thing' class object in panoptic segmentation through model ensemble.

\subsection{Result}
In the third PVUW Challenge, we rank first in the development phase and second in the test phase. The ranking lists for the development and test phases are shown in Table 1 and Table 2, respectively. Our method achieve VPQ of 56.36 and 57.12 respectively during the development and testing phases, demonstrating strong segmentation performance. In addition, our method has significant advantages in tracking performance. The qualitative results of the VIPSeg test set are shown in Figure 4,which demonstrate strong segmentation and tracking performance in handling stuff and thing objects. Compared to the baseline DVIS++, our method improve by 1.19 on the VPQ metric and 0.0113 on the STQ metric. 

\begin{table}
  \centering
  \resizebox{1.0\columnwidth}{!}{
  \begin{tabular}{ccccccc}
    \toprule
    Team & VPQ & VPQ1 & VPQ2 & VPQ4 & VPQ6 & STQ \\
    \midrule
    SiegeLion&	56.3598& 57.1408& 56.4636& 56.0302& 55.8046& 0.5252   \\
     kevin1234&	55.6940 & 56.4139& 55.8574& 55.3925& 55.1122& 0.5190   \\
    Reynard&	54.5464  & 55.2727& 54.6924& 54.2534 & 53.9672 & 0.5166    \\
    ipadvideo&	54.2571 & 54.9604& 54.4390& 53.9786 & 53.6504 & 0.5093   \\
    zhangtao-whu&	52.7673 & 53.3162& 52.9243& 52.5669 & 52.2618 & 0.5016   \\
    \bottomrule
  \end{tabular}
  }
  \caption{Ranking results of  leaderboard during the development phase.}
  \label{tab:example1}
\end{table}

\begin{table}
  \centering
  \resizebox{1.0\columnwidth}{!}{
  \begin{tabular}{ccccccc}
    \toprule
    Team & VPQ & VPQ1 & VPQ2 & VPQ4 & VPQ6 & STQ \\
    \midrule
    kevin1234&	58.2585& 59.1009& 58.5042& 57.9007& 57.5283& 0.5434   \\
     SiegeLion&	57.1188 & 58.2143& 57.4119& 56.6798& 56.1691& 0.5397   \\
    Reynard&	57.0114 & 57.8900& 57.2240& 56.6509& 56.2807 & 0.5343   \\
    ipadvideo&	28.3810 & 29.1165& 28.6770& 28.1028 & 27.6277 & 0.2630   \\
    JMCarrot&	22.1060 & 23.8061& 22.8334& 21.4910 & 20.2935 & 0.2603   \\
    \bottomrule
  \end{tabular}
  }
  \caption{Ranking result of  leaderboard during the test phase.}
  \label{tab:example1}
\end{table}

\begin{table}
  \centering
  \resizebox{1.0\columnwidth}{!}{
  \begin{tabular}{ccccccc}
    \toprule
    Method & VPQ & VPQ1 & VPQ2 & VPQ4 & VPQ6 & STQ \\
    \midrule
    Baseline&	55.9332 & 57.0035& 56.2178& 55.5001& 55.0114& 0.5284   \\
    Ensemble(VSS)&	57.1188 & 58.2143& 57.4119& 56.6798& 56.1691& 0.5397   \\
    \bottomrule
  \end{tabular}
  }
  \caption{Ablation study of our method.}
  \label{tab:example1}
\end{table}

\section{Conclusion}
\label{sec:Conclusion}

In this paper, we propose a robust solution for the task of video panoptic segmentation and make nontrivial improvements and attempts in many stages such as model, training and ensemble. In the end, we introduce DVIS++ to the VPS field and verify that the decoupling strategy proposed by DVIS++ significantly improves the performance for both thing and stuff objects. Then,we add an additional image semantic segmentation model to further improve the performance of semantic classes. As a result, we get the 2nd place in the VPS track of the PVUW Challenge 2024, scoring 56.36 VPQ and 57.12 VPQ in the development and test phases, respectively.


\begin{thebibliography}{23}

\bibitem[1]{rw0001}
Alexander Kirillov, Kaiming He, Ross Girshick, Carsten Rother, and Piotr Dollár.
\newblock Panoptic segmentation.
\newblock {\em In IEEE CVPR, pages 9396-9405, 2019.}

\bibitem[2]{rw0002}
Yanwei Li, Xinze Chen, Zheng Zhu, Lingxi Xie, Guan Huang, Dalong Du, and Xingang Wang.
\newblock Attention-guided unified network for panoptic segmentation.
\newblock {\em In IEEE CVPR, pages 7019-7028, 2019.}

\bibitem[3]{rw0003}
Huanyu Liu, Chao Peng, Changqian Yu, Jingbo Wang, Xu~Liu, Gang Yu, and Wei Jiang.
\newblock An end-to-end network for panoptic segmentation.
\newblock {\em In IEEE CVPR, pages 6165-6174, 2019.}

\bibitem[4]{rw0004}
Yuwen Xiong, Renjie Liao, Hengshuang Zhao, Rui Hu, Min Bai, Ersin Yumer, and Raquel Urtasun.
\newblock Upsnet: A unified panoptic segmentation network.
\newblock {\em In IEEE CVPR, pages 8810-8818, 2019.}

\bibitem[5]{rw0005}
Shuyang Sun, Weijun Wang, Qihang Yu, Andrew Howard, Philip Torr, and Liang-Chieh Chen.
\newblock Remax: Relaxing for better training on efficient panoptic segmentation.
\newblock {\em In arXiv preprint, arXiv:2306.17319, 2023.}

\bibitem[6]{vps2020}
Joon-Young~Lee Dahun~Kim, Sanghyun~Woo and In~So Kweon.
\newblock Video panoptic segmentation.
\newblock {\em In Proceedings of the IEEE/CVF Conference on Computer Vision and Pattern Recognition, pages 9859–9868, 2020.}

\bibitem[7]{rw0008}
Bowen Cheng, Maxwell~D. Collins, Yukun Zhu, Ting Liu, Thomas~S. Huang, Hartwig Adam, and Liang-Chieh Chen.
\newblock Panoptic-deeplab: A simple, strong, and fast baseline for bottom-up panoptic segmentation.
\newblock {\em In IEEE CVPR, pages 12472-12482, 2020.}

\bibitem[8]{rw0009}
Siyuan Qiao, Yukun Zhu, Hartwig Adam, Alan Yuille, and Liang-Chieh Chen.
\newblock Vip-deeplab: Learning visual perception with depth-aware video panoptic segmentation.
\newblock {\em In IEEE CVPR, pages 3996-4007, 2021.}

\bibitem[9]{rw0010}
Huiyu Wang, Yukun Zhu, Hartwig Adam, Alan Yuille, and Liang-Chieh Chen.
\newblock Max-deeplab: End-to-end panoptic segmentation with mask transformers.
\newblock {\em In IEEE CVPR, pages 5459-5470, 2021.}

\bibitem[10]{rw0011}
Yi~Zhou, Hui Zhang, Hana Lee, Shuyang Sun, Pingjun Li, Yangguang Zhu, ByungIn Yoo, Xiaojuan Qi, and Jae-Joon Han.
\newblock Slot-vps: Object-centric representation learning for video panoptic segmentation.
\newblock {\em In arXiv preprint, arXiv:2112.08949, 2021.}

\bibitem[11]{rw0012}
Tao Zhang, Xingye Tian, Yu~Wu, Shunping Ji, Xuebo Wang, Yuan Zhang, and Pengfei Wan.
\newblock Dvis: Decoupled video instance segmentation framework.
\newblock {\em In IEEE ICCV, pages 1282-1291, 2023.}

\bibitem[12]{rw0013}
Tao Zhang, Xingye Tian, Haoran Wei, Yu~Wu, Shunping Ji, Xuebo Wang, Xin Tao, Yuan Zhang, and Pengfei Wan.
\newblock 1st place solution for pvuw challenge 2023: Video panoptic segmentation.
\newblock {\em In arXiv preprint, arXiv:2306.04091, 2023.}

\bibitem[13]{zhang2023dvisplus}
Tao Zhang, Xingye Tian, Yikang Zhou, Shunping Ji, Xuebo Wang, Xin Tao, Yuan Zhang, Pengfei Wan, Zhongyuan Wang, and Yu~Wu.
\newblock Dvis++: Improved decoupled framework for universal video segmentation.
\newblock {\em arXiv preprint arXiv:2312.13305}, 2023.

\bibitem[14]{rw0015}
Inkyu Shin, Dahun Kim, Qihang Yu, Jun Xie, Hong-Seok Kim, Bradley Green, In~So Kweon, Kuk-Jin Yoon, and Liang-Chieh Chen.
\newblock Video-kmax: A simple unified approach for online and near-online video panoptic segmentation.
\newblock {\em In IEEE WACV, pages 228-238, 2024.}

\bibitem[15]{rw0016}
Ju~He, Qihang Yu, Inkyu Shin, Xueqing Deng, Xiaohui Shen, Alan Yuille, and Liang-Chieh Chen.
\newblock Maxtron: Mask transformer with trajectory attention for video panoptic segmentation.
\newblock {\em In arXiv preprint, arXiv:2311.18537, 2023.}

\bibitem[16]{chen2022vitadapter}
Zhe Chen, Yuchen Duan, Wenhai Wang, Junjun He, Tong Lu, Jifeng Dai, and Yu~Qiao.
\newblock Vision transformer adapter for dense predictions.
\newblock {\em arXiv preprint arXiv:2205.08534}, 2022.

\bibitem[17]{Dinov2}
T.~Moutakanni H. Vo M. Szafraniec V. Khalidov P. Fernandez D. Haziza F. Massa A. El-Nouby et~al. M.~Oquab, T.~Darcet.
\newblock Dinov2: Learning robust visual features without supervision.
\newblock {\em arXiv preprint arXiv:2304.07193, 2023.}

\bibitem[18]{cheng2022masked}
Bowen Cheng, Ishan Misra, Alexander~G Schwing, Alexander Kirillov, and Rohit Girdhar.
\newblock Masked-attention mask transformer for universal image segmentation.
\newblock In {\em Proceedings of the IEEE/CVF Conference on Computer Vision and Pattern Recognition}, pages 1290--1299, 2022.

\bibitem[19]{miao2022large}
Jiaxu Miao, Xiaohan Wang, Yu~Wu, Wei Li, Xu~Zhang, Yunchao Wei, and Yi~Yang.
\newblock Large-scale video panoptic segmentation in the wild: A benchmark.
\newblock In {\em Proceedings of the {IEEE} Conference on Computer Vision and Pattern Recognition}, 2022.

\bibitem[20]{miao2021vspw}
Jiaxu Miao, Yunchao Wei, Yu~Wu, Chen Liang, Guangrui Li, and Yi~Yang.
\newblock Vspw: A large-scale dataset for video scene parsing in the wild.
\newblock In {\em Proceedings of the IEEE/CVF conference on computer vision and pattern recognition}, pages 4133--4143, 2021.

\bibitem[21]{stq2021}
Maxwell Collins Yukun Zhu Paul Voigtlaender Hartwig Adam Bradley Green Andreas Geiger Bastian Leibe Daniel Cremers-et~al Mark~Weber, Jun~Xie.
\newblock Step: Segmenting and tracking every pixel.
\newblock {\em arXiv preprint arXiv:2102.11859, 2021.}

\bibitem[22]{pq2019}
Kaiming~He Alexander~Kirillov, Ross~Girshick and Piotr Dollar.
\newblock Panoptic feature pyramid networks.
\newblock {\em In ´IEEE CVPR, pages 6399–6408, 2019.}

\bibitem[23]{coco}
Serge Belongie James Hays Pietro Perona Deva Ramanan Piotr~Dollar Tsung-Yi~Lin, Michael~Maire and C~Lawrence~´ Zitnick.
\newblock Microsoft coco: Common objects in context.
\newblock {\em In Computer Vision–ECCV 2014: 13th European Conference, Zurich, Switzerland, September 6-12, 2014, Proceedings, Part V 13, pages 740–755. Springer, 2014.}

\end{thebibliography}
\end{document}